\documentclass[sigconf,nonacm=true]{acmart}
\usepackage[utf8]{inputenc}
\usepackage{cite}
\usepackage{amsmath,amssymb,amsfonts}
\usepackage{algorithmic}
\usepackage{bm}
\usepackage{algorithm}
\usepackage{graphicx}
\usepackage{textcomp}
\usepackage{xcolor}
\usepackage{listings}
\usepackage{xargs}
\usepackage{xcolor}
\usepackage{todonotes}
\usepackage{tabularx}
\usepackage{booktabs}
\usepackage{float}
\usepackage{enumitem}
\usepackage{setspace}
\usepackage{xspace}
\usepackage{fixfoot}
\newfloat{algorithm}{t}{lop}
\DeclareFixedFootnote*{\redacted}{Repository location redacted for blind review and will be added to final paper.}

\newcommandx{\todoitem}[2][1=]{\todo[linecolor=blue,backgroundcolor=blue!25,bordercolor=blue,#1]{#2}}

\copyrightyear{2019}
\acmYear{2019}
\setcopyright{none}
\acmConference[SC19]{SC19}{November 17--22, 2019}{Denver, CO}
\begin{document}

\title{Etalumis: Bringing Probabilistic Programming to Scientific Simulators at Scale}


\author{Atılım Güneş Baydin}
\affiliation{\institution{University of Oxford}}

\author{Lei Shao}
\affiliation{\institution{Intel Corporation}}

\author{Wahid Bhimji}
\affiliation{\institution{Lawrence Berkeley National Laboratory}}

\author{Lukas Heinrich}
\affiliation{\institution{CERN}}

\author{Lawrence Meadows}
\affiliation{\institution{Intel Corporation}}

\author{Jialin Liu}
\affiliation{\institution{Lawrence Berkeley National Laboratory}}

\author{Andreas Munk}
\affiliation{\institution{University of British Columbia}}

\author{Saeid Naderiparizi}
\affiliation{\institution{University of British Columbia}}

\author{Bradley Gram-Hansen}
\affiliation{\institution{University of Oxford}}

\author{Gilles Louppe}
\affiliation{\institution{University of Liège}}

\author{Mingfei Ma}
\affiliation{\institution{Intel Corporation}}

\author{Xiaohui Zhao}
\affiliation{\institution{Intel Corporation}}

\author{Philip Torr}
\affiliation{\institution{University of Oxford}}

\author{Victor Lee}
\affiliation{\institution{Intel Corporation}}

\author{Kyle Cranmer}
\affiliation{\institution{New York University}}

\author{Prabhat}
\affiliation{\institution{Lawrence Berkeley National Laboratory}}

\author{Frank Wood}
\affiliation{\institution{University of British Columbia}}

\renewcommand{\shortauthors}{Baydin, Shao, Bhimji, Heinrich, Meadows, Liu, Munk, Naderiparizi, Gram-Hansen, Louppe, Ma, Zhao, Torr, Lee, Cranmer, Prabhat, Wood}

\begin{abstract}
Probabilistic programming languages (PPLs) are receiving widespread attention for performing Bayesian inference in complex generative models. However, applications to science remain limited because of the impracticability of rewriting complex scientific simulators in a PPL, the computational cost of inference, and the lack of scalable implementations. To address these, we present a novel PPL framework that couples directly to existing scientific simulators through a cross-platform probabilistic execution protocol and provides Markov chain Monte Carlo (MCMC) and deep-learning-based inference compilation (IC) engines for tractable inference. To guide IC inference, we perform distributed training of a dynamic 3DCNN--LSTM architecture with a PyTorch-MPI-based framework on 1,024 32-core CPU nodes of the Cori supercomputer with a global minibatch size of 128k: achieving a performance of 450 Tflop/s through enhancements to PyTorch. We demonstrate a Large Hadron Collider (LHC) use-case with the C++ Sherpa simulator and achieve the largest-scale posterior inference in a Turing-complete PPL.

\end{abstract}
\maketitle

\section{Introduction}
Probabilistic programming \citep{vandemeent2018intro} is an emerging paradigm within machine learning that uses general-purpose programming languages to express probabilistic models. This is achieved by introducing statistical conditioning as a language construct so that inverse problems can be expressed. Probabilistic programming languages (PPLs) have semantics \citep{staton2016semantics} that can be understood as Bayesian inference \citep{ghahramani2015probabilistic,gelman2013bayesian,bishop2006pattern}. The major challenge in designing useful PPL systems is that language evaluators must solve arbitrary, user-provided inverse problems, which usually requires general-purpose inference algorithms that are computationally expensive.

In this paper we report our work that enables, for the first time, the use of \emph{existing} stochastic simulator code as a probabilistic program in which one can do fast, repeated (amortized) Bayesian inference; this enables one to predict the distribution of input parameters and all random choices in the simulator from an observation of its output. In other words, given a simulator of a generative process in the forward direction (inputs$\to$outputs), our technique can provide the reverse (outputs$\to$inputs) by predicting the whole latent state of the simulator that could have given rise to an observed instance of its output. For example, using a particle physics simulation we can get distributions over the particle properties and decays within the simulator that can give rise to a collision event observed in a detector, or, using a spectroscopy simulator we can determine the elemental matter composition and dispersions within the simulator explaining an observed spectrum. In fields where accurate simulators of real-world phenomena exist, our technique enables the interpretable explanation of real observations under the structured model defined by the simulator code base. 

We achieve this by defining a probabilistic programming execution protocol that interfaces with existing simulators at the sites of random number draws, without altering the simulator's structure and execution in the host system. The random number draws are routed through the protocol to a PPL system which treats these as samples from corresponding prior distributions in a Bayesian setting, giving one the capability to record or guide the execution of the simulator to perform inference. Thus we generalize existing simulators as probabilistic programs and make them subject to inference under general-purpose inference engines.

Inference in the probabilistic programming setting is performed by sampling in the space of execution traces, where a single sample (an execution trace) represents a full run of the simulator. Each execution trace itself is composed of a potentially unbounded sequence of addresses, prior distributions, and sampled values, where an address is a unique label identifying each random number draw. In other words, we work with empirical distributions over simulator executions, which entails unique requirements on memory, storage, and computation that we address in our implementation. The addresses comprising each trace give our technique the unique ability to provide direct connections to the simulator code base for any predictions at test time, where the simulator is no longer used as a black box but as a highly structured and interpretable probabilistic generative model that it implicitly represents.

Our PPL provides inference engines from the Markov chain Monte Carlo (MCMC) and importance sampling (IS) families. MCMC inference guarantees closely approximating the true posterior of the simulator, albeit with significant computational cost due to its sequential nature and the large number of iterations one needs to accumulate statistically independent samples. Inference compilation (IC) \citep{le2016inference} addresses this by training a dynamic neural network to provide proposals for IS, leading to fast amortized inference.

We name this project ``Etalumis'', the word ``simulate'' spelled backwards, as a reference to the fact that our technique essentially inverts a simulator by probabilistically inferring all choices in the simulator given an observation of its output. We demonstrate this by inferring properties of particles produced at the Large Hadron Collider (LHC) using the Sherpa\footnote{https://gitlab.com/sherpa-team/sherpa}  \citep{gleisberg2009event} simulator.

\subsection{Contributions}
Our main contributions are:
\begin{itemize}[leftmargin=*]
	\item A novel PPL framework that enables execution of existing stochastic simulators under the control of general-purpose inference engines, with HPC features including handling multi-TB data  and distributed training and inference.
	\item The largest scale posterior inference in a Turing-complete PPL, where our experiments encountered approximately 25,000 latent variables\footnote{Note that the simulator defines an unlimited number of random variables because of the presence of rejection sampling loops.} expressed by the existing Sherpa simulator code base of nearly one million lines of code in C++ \citep{gleisberg2009event}.
    \item Synchronous data parallel training of a dynamic 3DCNN--LSTM neural network (NN) architecture using the PyTorch \citep{paszke2017automatic} MPI framework at the scale of 1,024 nodes (32,768 CPU cores) with a global minibatch size of 128k. To our knowledge this is the largest scale use of PyTorch's builtin MPI functionality,\footnote{Personal communication with PyTorch developers.} and the largest minibatch size used for this form of NN model.
\end{itemize}

\section{Probabilistic programming for particle physics}
\label{sec:probprog_physics}
Particle physics seeks to understand particles produced in collisions at accelerators such at the LHC at CERN. Collisions happen millions of times per second, creating cascading particle decays, observed in complex instruments such as the ATLAS detector \citep{atlas}, comprising millions of electronics channels. These experiments analyze the vast volume of resulting data and seek to reconstruct the initial particles produced in order to make discoveries including physics beyond the current Standard Model of particle physics \citep{glashow1961partial}\citep{weinberg1967weinberg}\citep{salam1968proceedings}\citep{veltman1972regularization}.

The Standard Model has a number of parameters (e.g., particle masses), which we can denote $\mathbf{\theta}$, describing the way particles and fundamental forces act in the universe. In a given collision at the LHC, with initial conditions denoted $E$, we observe a cascade of particles interact with particle detectors. If we denote \emph{all} of the random ``choices'' made by nature as $\mathbf{x}$, the Standard Model describes, generatively, the conditional probability $p(\mathbf{x}\vert E,\mathbf{\theta})$, that is, the distribution of all choices $\mathbf{x}$ as a function of initial conditions $E$ and model parameters $\mathbf{\theta}$. Note that, while the Standard Model can be expressed symbolically in mathematical notation \citep{griffiths2008introduction,peskin2018introduction}, it can also be expressed computationally as a stochastic simulator \citep{gleisberg2009event}, which, given access to a random number generator, can draw samples from $p(\mathbf{x})$.\footnote{Dropping the dependence on $E$ and $\mathbf{\theta}$ because everything in this example is conditionally dependent on these quantities.} Similarly, a particle detector can be modeled as a stochastic simulator, generating samples from $p(\mathbf{y}\vert\mathbf{x})$, the likelihood of observation $\mathbf{y}$ as a function of $\mathbf{x}$.

In this paper we focus on a real use-case in particle physics, performing experiments on the decay of the $\tau$ (tau) lepton. This is under active investigation by LHC physicists \citep{tau} and important to uncovering properties of the Higgs boson. We use the state-of-the-art Sherpa simulator \citep{gleisberg2009event} for modeling $\tau$ particle creation in LHC collisions and their subsequent decay into further particles (the stochastic events $\mathbf{x}$ above), coupled to a fast 3D detector simulator for the detector observation $\mathbf{y}$.

Current methods in the field include performing classification and regression using machine learning approaches on low dimensional distributions of derived variables \citep{tau} that provide point-estimates without the posterior of the full latent state nor the deep interpretability of our approach. Inference of the latent structure has only previously been used in the field with drastically simplified models of the process and detector \citep{kondo1988dynamical}  \citep{atlasmem}.

PPLs allow us to express inference problems such as: given an actual particle detector observation $\mathbf{y}$, what sequence of choices $\mathbf{x}$ are likely to have led to this observation? In other words, we would like to find $p(\mathbf{x}\vert\mathbf{y})$, the distribution of $\mathbf{x}$ as a function of $\mathbf{y}$. To solve this inverse problem via conditioning requires invoking Bayes rule 
\begin{align*}
p(\mathbf{x}\vert\mathbf{y}) = \frac{p(\mathbf{y},\mathbf{x})}{p(\mathbf{y})} =
\frac{p(\mathbf{y}\vert\mathbf{x})p(\mathbf{x})}{\int p(\mathbf{y}\vert\mathbf{x})p(\mathbf{x}) d\mathbf{x}}
\end{align*}
where the posterior distribution of interest, $p(\mathbf{x}|\mathbf{y})$, is related to the composition of the two stochastic simulators in the form of the joint distribution $p(\mathbf{y},\mathbf{x})=p(\mathbf{y}|\mathbf{x})p(\mathbf{x})$ renormalized by the marginal probability, or evidence of the data, $p(\mathbf{y}) = \int p(\mathbf{y}|\mathbf{x})p(\mathbf{x}) d\mathbf{x}.$  Computing the evidence requires summing over all possible paths that the simulation can take. This is a large number of possible paths; in most models this is a quantity that is impossible to compute in polynomial time. In practice PPLs approximate the posterior $p(\mathbf{x}\vert\mathbf{y})$ using sampling-based inference engines that sidestep the integration problem but remain computationally intensive. This specifically is where probabilistic programming meets, for the first time in this paper, high-performance computing.

\section{State of the art}
\subsection{Probabilistic programming}
Within probabilistic programming, recent advances in computational hardware have made it possible to distribute certain types of inference processes, enabling inference to be applied to problems of real-world relevance~\citep{tran2016edward}. By parallelizing computation over several cores, PPLs have been able to perform large-scale inference on models with increasing numbers of observations, such as the cause and effect analysis of $1.6\times 10^9$  genetic measurements~\citep{gopalan2016scaling, tran2016edward}, spatial analysis of $1.5\times 10^4$ shots from 308 NBA players~\citep{dieng2017variational}, exploratory analysis of $1.7 \times 10^6$ taxi trajectories~\citep{hoffman2013stochastic}, and probabilistic modeling for processing hundreds-of-thousands of Xbox live games per day to rank and match players fairly \citep{herbrich2007trueskill,InferNET18}.

In all these large-scale programs, despite the number of observations being large, model sizes in terms of the number of latent variables have been limited \citep{hoffman2013stochastic}. In contrast, to perform inference in a complex scientific model such as the Standard Model encoded by Sherpa requires handling thousands of latent variables, all of which need to be controlled within the program to perform inference in a scalable manner. To our knowledge, no existing PPL system has been used to run inference at the scale we are reporting in this work, and instances of distributed inference in existing literature have been typically restricted to small clusters~\citep{dillon2017tensorflow}.

A key feature of PPLs is that they decouple model specification from inference. A model is implemented by the user as a stand-alone regular program in the host programming language, specifying a generative process that produces samples from the joint prior distribution $p(\mathbf{y},\mathbf{x})=p(\mathbf{y}\vert\mathbf{x})p(\mathbf{x})$ in each execution, that is, a forward model going from choices $\mathbf{x}$ to outcomes (observations) $\mathbf{y}$. The same program can then be executed using a variety of general-purpose inference engines available in the PPL system to obtain $p(\mathbf{x}\vert\mathbf{y})$, the inverse going from observations $\mathbf{y}$ to choices $\mathbf{x}$. Inference engines available in PPLs range from MCMC-based lightweight Metropolis Hastings (LMH) \citep{wingate2011lightweight} and random-walk Metropolis Hastings (RMH) \citep{le2015rmh} algorithms to importance sampling (IS) \citep{arulampalam2002tutorial} and sequential Monte Carlo \citep{doucet2009tutorial}. Modern PPLs such as Pyro \citep{bingham2018pyro} and TensorFlow Probability \citep{tran2016edward,dillon2017tensorflow} use gradient-based inference engines including variational inference \citep{kingma2013auto,hoffman2013stochastic} and Hamiltonian Monte Carlo \citep{neal2011mcmc,hoffman2014no} that benefit from modern deep learning hardware and automatic differentiation \citep{baydin2018automatic} features provided by PyTorch \citep{paszke2017automatic} and TensorFlow \citep{abadi2016tensorflow} libraries. Another way of making use of gradient-based optimization is to combine IS with deep-learning-based proposals trained with data sampled from the probabilistic program, resulting in the IC algorithm \citep{le2016inference,lezcano2017improvements} in an amortized inference setting \citep{gershman2014amortized}.

\subsection{Distributed training for deep learning}
To perform IC inference in Turing-complete PPLs in general, we would like to support the training of dynamic NNs whose runtime structure changes in each execution of the probabilistic model by rearranging NN modules corresponding to different addresses (unique random number draws) encountered \citep{le2016inference} (Section~\ref{sec:nn}). Moreover, depending on probabilistic model complexity, the NNs may grow in size if trained in an online setting, as a model can represent a potentially unbounded number of random number draws. In addition to these, the volume of training data required is large, as the data keeps track of all execution paths within the simulator. To enable rapid prototyping, model evaluation, and making use of HPC capacity, scaling deep learning training to multiple computation units is highly desirable \citep{Amrita:2018:EDL:3291743,Kurth:2018:EDL:3291656.3291724,kurth2017deep,Iandola2015,mathuriya2017scaling}.

In this context there are three prominent parallelism strategies: data- and model-parallelism, and layer pipelining. In this project we work in a data-parallel setting where different nodes train the same model on different subsets of data. For such training, there are synchronous- and asynchronous-update approaches. In synchronous update \citep{Pan2017,Das2016}, locally computed gradients are summed across the nodes at the same time with synchronization barriers for parameter update. In asynchronous update \citep{Zheng2016,Dean2012,Niu2011}, one removes the barrier so that nodes can independently contribute to a parameter server. Although synchronous update can entail challenges due to straggler effects \citep{chen2016revisiting,teng2018bayesian}, it has desirable properties in terms of convergence, reproducibility, and ease of debugging. In this work, given the novelty of the probabilistic techniques we are introducing and the need to fully understand and compare trained NNs without ambiguity, we employ synchronous updates.

In synchronous updates, large global minibatches can make convergence challenging and hinder test accuracy. \citet{lb_training} pointed out large-minibatch training can lead to sharp minima and a generalization gap. Other work \citep{Goyal,lars-paper} argues that the difficulties in large-minibatch training are optimization related and can be mitigated with learning rate scaling \citep{Goyal}. \citet{lars-paper} apply layer-wise adaptive rate scaling (LARS) to achieve large-minibatch-size training of a Resnet-50 architecture without loss of accuracy, and \citet{larc-paper} use layer-wise adaptive rate control (LARC) to improve training stability and speed. \citet{dont_decay_LR} have proposed to increase the minibatch size instead of decaying the learning rate, and more recent work \citep{critical_batch_size,batch_size_vs_train_steps} showed relationships between gradient noise scale (or training steps) and minibatch size. Through such methods, distributed training has been scaled to many thousands of CPUs or GPUs \citep{Amrita:2018:EDL:3291743,kurth2017deep, Kurth:2018:EDL:3291656.3291724, DBLP:journals/corr/abs-1811-05233}. While we take inspiration from these recent approaches, our dynamic NN architecture and training data create a distinct training setting which requires appropriate innovations, as discussed in Section~\ref{sec:nn}.

\section{Innovations}

\begin{figure}
    \centering
    \includegraphics[width=0.49\textwidth]{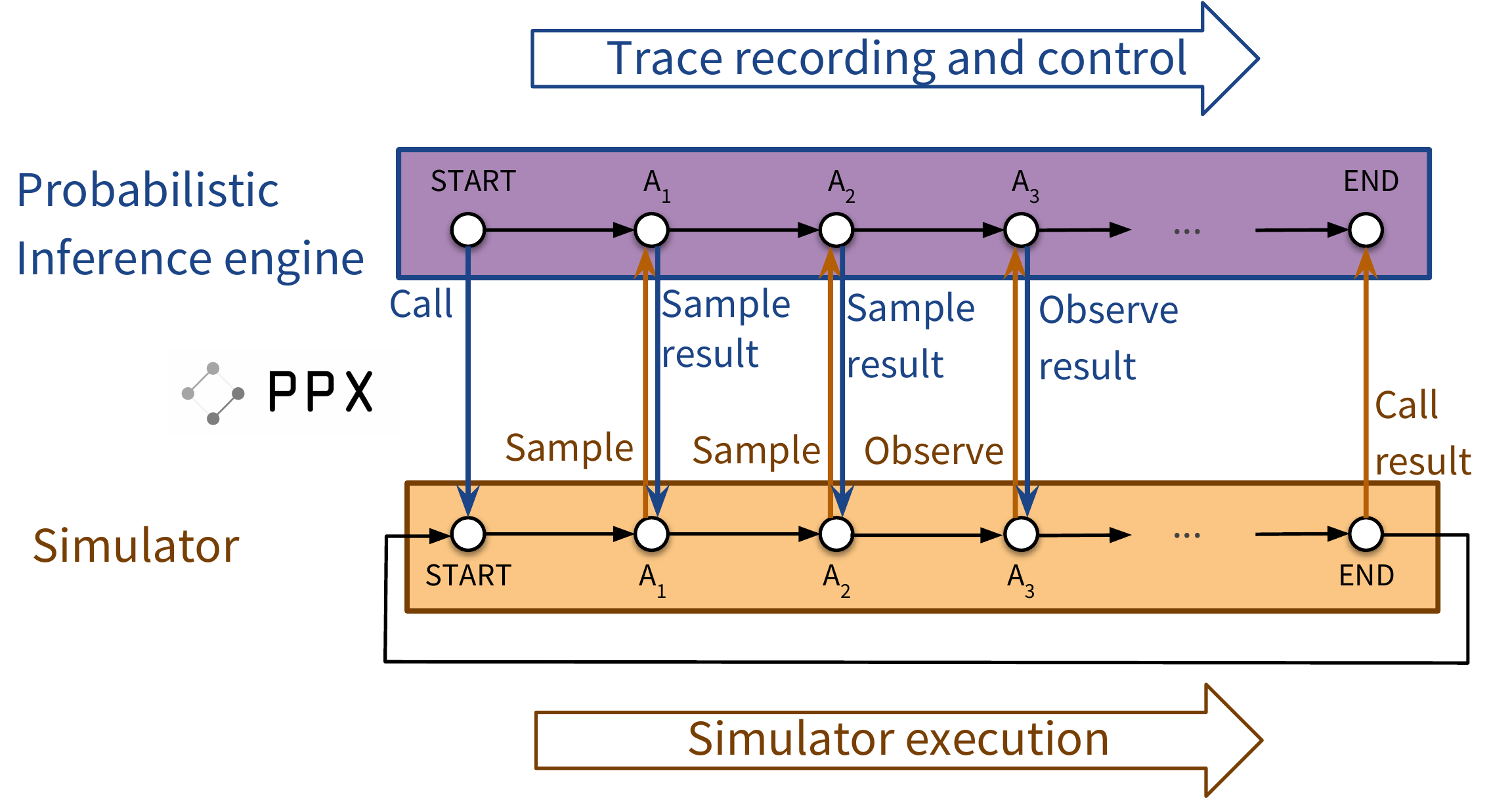}
    \caption{The probabilistic execution protocol (PPX). \emph{Sample} and \emph{observe} statements correspond to random number draws and conditioning, respectively.}
    \label{fig:ppx}
\end{figure}

\subsection{PPX and pyprob: executing existing simulators as probabilistic programs}
\label{sec:pyprob}
One of our main contributions in Etalumis is the development of a probabilistic programming execution protocol (PPX), which defines a cross-platform API for the execution and control of stochastic simulators\footnote{\url{https://github.com/probprog/ppx}} (Figure~\ref{fig:ppx}). The protocol provides language-agnostic definitions of common probability distributions and message pairs covering the call and return values of: (1) program entry points; (2) \emph{sample} statements for random number draws; and (3) \emph{observe} statements for conditioning. The purpose of this protocol is twofold:

\begin{itemize}[leftmargin=*]
\item It allows us to record execution traces of a stochastic simulator as a sequence of \emph{sample} and \emph{observe} (conditioning) operations on random numbers, each associated with an address $A_t$. We can use these traces for tasks such as inspecting the probabilistic model implemented by the simulator, computing likelihoods, learning surrogate models, and generating training data for IC NNs.
\item It allows us to control the execution of the simulator, at inference time, by making intelligent choices for each random number draw as the simulator keeps requesting random numbers. General-purpose PPL inference guides the simulator by making random number draws not from the prior $p(\mathbf{x})$ but from proposal distributions $q(\mathbf{x}\vert\mathbf{y})$ that depend on observed data $\mathbf{y}$ (Section~\ref{sec:probprog_physics}).
\end{itemize}

PPX is based on flatbuffers,\footnote{\url{http://google.github.io/flatbuffers/}} a streamlined version of Google protocol buffers, providing bindings into C++, C\#, Go, Java, JavaScript, PHP, Python, and TypeScript, enabling lightweight PPL front ends in these languages---in the sense of requiring the implementation of a simple intermediate layer to perform \emph{sample} and \emph{observe} operations over the protocol. We exchange PPX messages over ZeroMQ\footnote{\url{http://zeromq.org/}} \citep{hintjens2013zeromq} sockets, which allow communication between separate processes in the same machine (via inter-process sockets) or across a network (via TCP). PPX is inspired by the Open Neural Network Exchange (ONNX) project\footnote{https://onnx.ai/} allowing interoperability between major deep learning frameworks, and it allows the execution of any stochastic simulator under the control of any PPL system, provided that the necessary bindings are incorporated on both sides.

Using the PPX protocol as the interface, we implement two main components: (1) pyprob, a PyTorch-based PPL\footnote{\url{https://github.com/probprog/pyprob}} in Python and (2) a C++ binding to the protocol to route the random number draws in Sherpa to the PPL and therefore allow probabilistic inference in this simulator. Our PPL is designed to work with models written in Python and other languages supported through PPX. This is in contrast to existing PPLs such as Pyro \citep{bingham2018pyro} and TensorFlow Probability \citep{tran2016edward,dillon2017tensorflow} which do not provide a way to interface with existing simulators and require one to implement any model from scratch in the specific PPL.\footnote{We are planning to provide PPX bindings for these PPLs in future work.} We develop pyprob based on PyTorch \citep{paszke2017automatic}, to utilize its automatic differentiation \citep{baydin2018automatic} infrastructure with support for dynamic computation graphs for IC inference.

\subsection{Efficient Bayesian inference}
\label{sec:efficient_bayesian_inference}

Working with existing simulators as probabilistic programs restricts the class of inference engines that we can put to use. Modern PPLs commonly use gradient-based inference such as Hamiltonian Monte Carlo \citep{neal2011mcmc} and variational inference \citep{hoffman2013stochastic,kingma2013auto} to approximate posterior distributions. However this is not applicable in our setting due to the absence of derivatives in general simulator codes. Therefore in pyprob we focus our attention on two inference engine families that can control Turing-complete simulators over the PPX protocol: MCMC in the RMH variety \citep{wingate2011lightweight,le2015rmh}, which provides a high-compute-cost sequential algorithm with statistical guarantees to closely approximate the posterior, and IS with IC \citep{le2016inference}, which does not require derivatives of the simulator code but still benefits from gradient-based methods by training proposal NNs and using these to significantly speed up IS inference.

It is important to note that the inference engines in pyprob work in the space of execution traces of probabilistic programs, such that a single sample from the inference engine corresponds to a full run of the simulator. Inference in this setting amounts to making adjustments to the random number draws, re-executing the simulator, and scoring the resulting execution in terms of the likelihood of the given observation. Depending on the specific observation and the simulator code involved, inference is computationally very expensive, requiring up to millions of executions in the RMH engine. Despite being very costly, RMH provides a way of sampling from the true posterior \citep{neal1993probabilistic,neal2011mcmc}, which is needed in initial explorations of any new simulator to establish correct posteriors serving as reference to confirm that IC inference can work correctly in the given setting. To establish the correctness of our inference results, we implement several MCMC convergence diagnostics. Autocorrelation measures the number of iterations one needs to get effectively independent samples in the same MCMC chain, which allows us to estimate how long RMH needs to run to reach a target effective sample size. The Gelman--Rubin metric, given multiple independent MCMC chains sampled from the same posterior, compares the variance of each chain to the pooled variance of all chains to statistically establish that we converged on the true posterior \citep{gelman2013bayesian}.

RMH comes with a high computational cost. This is because it requires a large number of initial samples to be generated that are then discarded, of the order $\sim10^6$ for the Sherpa model we present in this paper. This is required to find the posterior density, which, as the model begins from an arbitrary point of the prior, can be very far from the starting region. Once this ``burn-in'' stage is completed the MCMC chain should be sampling from within the region containing the posterior. In addition to this, the sequential nature of each chain limits our ability to parallelize the computation, again creating computational inefficiencies in the high-dimensional space of simulator execution traces that we work with in our technique.

In order to provide fast, repeated inference in a distributed setting, we implement the IC algorithm, which trains a deep recurrent NN to provide proposals for an IS scheme \citep{le2016inference}. This works by running the simulator many times and therefore sampling a large set of execution traces from the simulator prior $p(\mathbf{x},\mathbf{y})$, and using these to train a NN that represents $q(\mathbf{x}\vert\mathbf{y})$, i.e., informed proposals for random number draws $\mathbf{x}$ given observations $\mathbf{y}$, by optimizing the loss $\mathcal{L}(\mathbf{\phi})=\mathbb{E}_{p(\mathbf{y})}\left[D_{\mathrm{KL}}(p(\mathbf{x}\vert\mathbf{y})\vert\vert q_{\mathbf{\phi}}(\mathbf{x}\vert\mathbf{y}))\right] = \mathbb{E}_{p(\mathbf{x},\mathbf{y})}\left[-\log q_{\mathbf{\phi}}(\mathbf{x}\vert\mathbf{y})\right] + \mathrm{const.}$, where $\mathbf{\phi}$ are NN parameters (Algorithm~\ref{alg:minibatch_loss} and Figure~\ref{fig:nn}) \citep{le2016inference}. This phase of sampling the training data and training the NN is costly, but it needs to be performed only once for any given model. Once the proposal NN is trained to convergence, the IC inference engine becomes competitive in performance, which allows us to achieve a given effective sample size in the posterior $p(\mathbf{x}\vert\mathbf{y})$ using a fraction of the RMH computational cost. IC inference is embarrassingly parallel, where many instances of the same trained NN can be executed to run distributed inference on a given observation.

To further improve inference performance, we make several low-level improvements in the code base. The C++ front end of PPX uses concatenated stack frames of each random number draw as a unique address identifying a latent variable in the corresponding PPL model. Stack traces are obtained with the {\tt backtrace(3)} function as instruction addresses and then converted to symbolic names using the {\tt dladdr(3)} function \citep{linuxman2019}. The conversion is quite expensive, which prompted us to add a hash map to cache {\tt dladdr} results, giving a 5x improvement in the production of address strings that are essential in our inference engines. The particle detector simulator that we use was initially coded to use the xtensor library\footnote{\url{https://xtensor.readthedocs.io}} to implement the probability density function (PDF) of multivariate normal distributions in the general case, but was exclusively called on 3D data. This code was replaced by a scalar-based implementation limited to the 3D case, resulting in a 13x speed-up in the PDF, and a 1.5x speed-up of our simulator pipeline in general. The bulk of our further optimizations focus on the NN training for IC inference and are discussed in the next sections.

\subsection{Dynamic neural network architecture}
\label{sec:nn}
\begin{figure}
    \centering
    \includegraphics[width=0.49\textwidth]{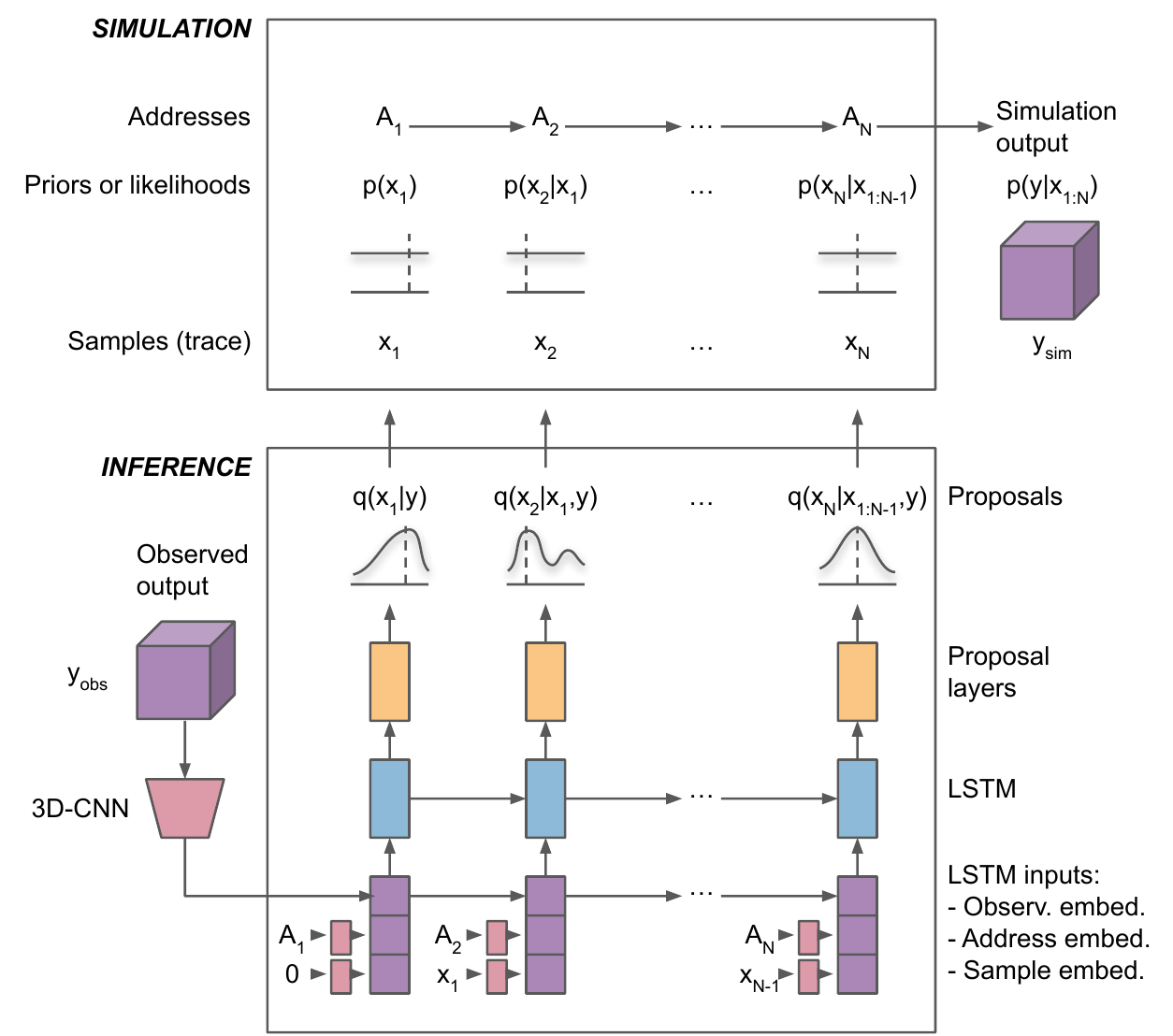}
    \caption{Simulation and inference. \emph{Top:} model addresses, priors and samples. \emph{Bottom:} IC inference engine proposals and NN architecture.}
    \label{fig:nn}
\end{figure}

The NN architecture used in IC inference is based on a LSTM \citep{hochreiter1997long} recurrent core that gets executed as many time steps as the simulator's probabilistic trace length (Figure~\ref{fig:nn}). To this core NN, various other NN components get attached according to the series of addresses $A_t$ executed in the simulator. In other words, we construct a dynamic NN whose runtime structure changes in each execution trace, implemented using the dynamic computation graph infrastructure in PyTorch. The input to this LSTM in each time step is a concatenation of embeddings of the observation, the current address in the simulator, and the previously sampled value. The observation embedding is a NN specific to the observation domain. Address embeddings are learned vectors representing the identity of random choices $A_t$ in the simulator address space. Sample embeddings are address-specific layers encoding the value of the random draw in the previous time step. The LSTM output, at each time step, is fed into address-specific proposal layers that provide the final output of the NN for IC inference: proposal distributions $q(\mathbf{x}\vert\mathbf{y})$ to use for each address $A_t$ as the simulator keeps running and requesting new random numbers over the PPX protocol (Section~\ref{sec:pyprob}).

\begin{figure}
    \centering
    \includegraphics[trim={30mm 10mm 16mm 16mm},clip,width=0.5\textwidth]{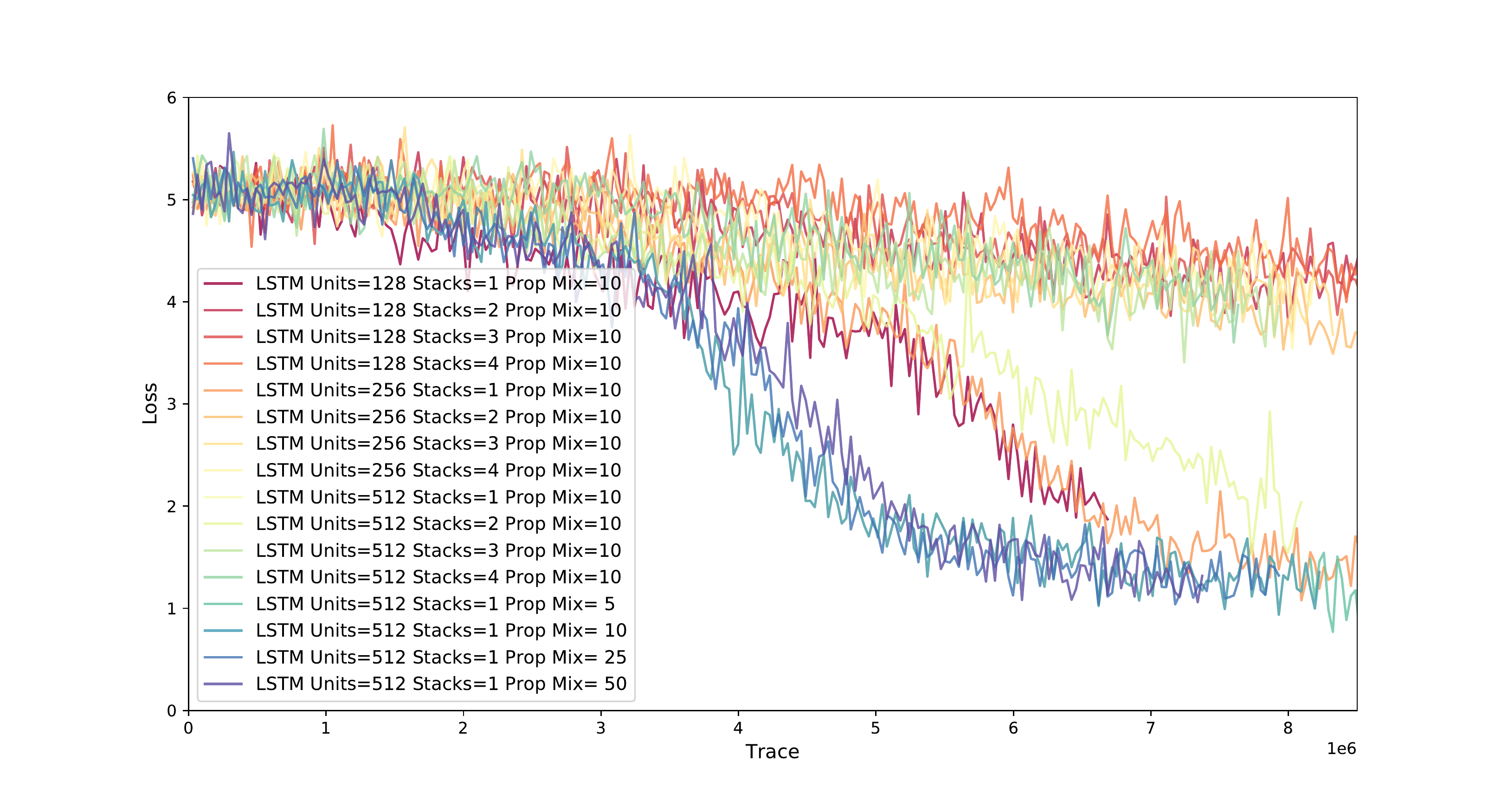}
    \caption{Loss curves for NN architectures considered in the hyperparameter search detailed in the text.}
    \label{fig:archhpo}
\end{figure}

For the Sherpa experiments reported in this paper, we work with 3D observations of size 35x35x20, representing particle detector voxels. To tune NN architecture hyperparameters, we search a grid of LSTM stacks in range \{1, 4\}, LSTM hidden units in the set \{128, 256, 512\}, and number of proposal mixture components in the set \{5, 10, 25, 50\} (Figure~\ref{fig:archhpo}). We settle on the following architecture: an LSTM with 512 hidden units; an observation embedding of size 256, encoded with a 3D convolutional neural network (CNN) \citep{lecun1998gradient} acting as a feature extractor, with layer configuration Conv3D(1, 64, 3)--Conv3D(64, 64, 3)--MaxPool3D(2)--Conv3D(64, 128, 3)--Conv3D(128, 128, 3)--Conv3D(128, 128, 3)-- MaxPool3D(2)--FC(2048, 256); previous sample embeddings of size 4 given by single-layer NNs; and address embeddings of size 64. The proposal layers are two-layer NNs, the output of which are either a mixture of ten truncated normal distributions \citep{bishop1994mixture} (for uniform continuous priors) or a categorical distribution (for categorical priors). We use ReLU nonlinearities in all NN components. All of these NN components except the LSTM and the 3DCNN are dependent on addresses $A_t$ in the simulator, and these address-specific layers are created at the first encounter with a random number draw at a given address. Thus the number of trainable parameters in an IC NN is dependent on the size of the training data, because the more data gets used, the more likely it becomes to encounter new addresses in the simulator.

The pyprob framework is capable of operating in an ``online'' fashion, where NN training and layer generation happens using traces sampled by executing the simulator on-the-fly and discarding traces after each minibatch, or ``offline'', where traces are sampled from the simulator and saved to disk as a dataset for further reuse (Algorithm~\ref{alg:distributed_training}). In our experiments, we used training datasets of 3M and 15M traces, resulting in NN sizes of 156,960,440 and 171,732,688 parameters respectively. All timing and scaling results presented in Sections \ref{sec:singlenoderesults} and \ref{sec:scalingperf} are performed with the larger network.

\subsection{Training of dynamic neural networks}
\label{sec:scaling_nn}
Scalable training of dynamic NNs we introduced in Section~\ref{sec:nn} pose unique challenges. Because of the address-dependent nature of the embedding and proposal layers of the overall IC NN, different nodes/ranks in a distributed training setting will work with different NN configurations according to the minibatch of training data they process at any given time. When the same NN is not shared across all nodes/ranks, it is not possible to rely on a generic allreduce operation for gradient averaging which is required for multi-node synchronous SGD. Inspired by neural machine translation (NMT) \citep{GNMT}, in the offline training mode with training data saved on the disk, we implemented the option of pre-processing the whole dataset to pre-generate all embedding and proposal layers that a given dataset would imply to exist. Once layer pre-generation is done, the collection of all embedding and proposal layers are shared on each node/rank. In this way, for offline training, we have a globally shared NN representing the superset of all NN components each node needs to handle in any given minibatch, thus making it possible to scale training of the NN on multiple nodes. 

Our allreduce-based training algorithm can also work in the online training setting, where training data is sampled from the simulator on-the-fly, if we freeze a globally shared NN and discard any subsequently encountered traces that contain addresses unknown at the time of NN architecture freezing. In future work, we intend to add a distributed open-ended implementation for online training to allow running without discarding, that will require the NN instances in each node/rank to grow with newly seen addresses.

\subsubsection{Single node improvements to Etalumis}
\label{sec:single_node_improvements_etalumis}
We profiled the Etalumis architecture with vtune, Cprofiler, and PyTorch autograd profiler, identifying data loading and 3D convolution as the primary computational hot-spots on which we focused our optimization efforts. We provide details on data loading and 3D convolution in the subsequent sections. In addition to these, execution traces from the Sherpa simulator have many different trace types  (a unique sequence of addresses $A_t$, with different sampled values) with different rates of occurrence: in a given dataset, some trace types can be encountered thousands of times while others are seen only once. This is problematic because at training time we further divide each minibatch into ``sub-minibatches'' based on trace type, where each sub-minibatch can be processed by the NN in a single forward execution due to all traces being of the same type, i.e., sharing the same sequence of addresses $A_t$ and therefore requiring the same NN structure (Algorithm \ref{alg:minibatch_loss}). Therefore minibatches containing more than one trace type do not allow for effective parallelization and vectorization. In other words, unlike conventional NNs, the effective minibatch size is determined by the average size of sub-minibatches, and the more trace types we have within a minibatch, the slower the computation. To address this, we explored multiple methods to enlarge effective minibatch size, such as sorting traces, multi-bucketing, and selectively batching traces from the same trace type together in each minibatch. These options and their trade offs are described in more detail in Section \ref{sec:discussion}.

\begin{algorithm}
\setstretch{0.75}
\caption{Computing minibatch loss $\mathcal{L}_n$ of NN parameters $\mathbf{\phi}$}
\label{alg:minibatch_loss}
\begin{algorithmic}
\REQUIRE Minibatch $\mathcal{D}_n$
\STATE $L\gets$ number of unique trace types found in $\mathcal{D}_n$
\STATE Construct sub-minibatches $\mathcal{D}^l_n$, for $l=1,\dots,L$
\STATE $\mathcal{L}_n\gets 0$
\FOR{$l\in\{1,\dots,L\}$}
\STATE{$\mathcal{L}_n\gets \mathcal{L}_n - \sum_{(\bm{x},\bm{y})\in\mathcal{D}^l_n}\log q_{\mathbf{\phi}}(\bm{x}|\bm{y})$}
\ENDFOR
\RETURN $\mathcal{L}_n$
\end{algorithmic}
\end{algorithm}

\begin{algorithm}
\setstretch{0.75}
\caption{Distributed training with MPI backend. $p(\bm{x},\bm{y})$ is the simulator and $\hat{G}(\bm{x},\bm{y})$ is an offline dataset sampled from $p(\bm{x},\bm{y})$}
\label{alg:distributed_training}
\begin{algorithmic}
\REQUIRE OnlineData \COMMENT{True/False value}
\REQUIRE $B$ \COMMENT{Minibatch size}
\STATE Initialize inference network $q_{\mathbf{\phi}}(\bm{x}|\bm{y})$
\STATE $N\gets $ number of processes
\FORALL{$n \in \{1,\dots,N\}$}
\WHILE{Not Stop}
\IF{OnlineData}
\STATE{Sample $\mathcal{D}_n=\{(\bm{x},\bm{y})_1,\dots,(\bm{x},\bm{y})_B\}$ from $p(\bm{x},\bm{y})$}
\ELSE
\STATE{Get $\mathcal{D}_n=\{(\bm{x},\bm{y})_1,\dots,(\bm{x},\bm{y})_B\}$ from $\hat{G}(\bm{x},\bm{y})$}
\ENDIF
\STATE Synchronize parameters ($\mathbf{\phi}$) across all processes 
\STATE $\mathcal{L}_n\gets -\frac{1}{B}\sum_{(\bm{x},\bm{y})\in\mathcal{D}_n}\log q_{\mathbf{\phi}}(\bm{x}|\bm{y})$
\STATE Calculate $\nabla_{\mathbf{\phi}}\mathcal{L}_n$
\STATE Call \texttt{all\_reduce} s.t. $\nabla_{\mathbf{\phi}} \mathcal{L}\gets\frac{1}{N}\sum_{n=1}^N \nabla_{\mathbf{\phi}}\mathcal{L}_n$
\STATE Update $\mathbf{\phi}$ using $\nabla_{\mathbf{\phi}} \mathcal{L}$ with e.g. ADAM, SGD, LARC, etc.
\ENDWHILE
\ENDFOR
\end{algorithmic}
\end{algorithm}

\subsubsection{Single node improvements to PyTorch}
\label{sec:singlenodeimprove}
The flexibility of dynamic computation graphs and competitive speed of PyTorch have been crucial for this project. Optimizations were performed on code belonging to Pytorch stable release v1.0.0 to better support this project on Intel\textsuperscript{\textregistered} Xeon\textsuperscript{\textregistered} CPU platforms, focused in particular on 3D convolution operations making use of the MKL-DNN open source math library. MKL-DNN uses a direct convolution algorithm and for a 5-dimensional input tensor with layout \{N, C, D, H, W\}, it is reordered into a layout of \{N, C, D, H, W, 8c\} which is more amenable for SIMD vectorization.\footnote{\url{https://intel.github.io/mkl-dnn/understanding_memory_formats.html}} The 3D convolution operator is vectorized on the innermost dimension which matches the 256-bit instruction length on AVX2, and parallelized on the outer dimensions. We also performed cache optimization to further improve performance. With these improvements we found the heavily used 3D convolution kernel achieved an 8x improvement on the Cori HSW platform.\footnote{\url{https://docs.nersc.gov/analytics/machinelearning/benchmarks/}}
The overall improvement on single node training time is given in Section \ref{sec:singlenoderesults}. These improvements are made available in a fork of the official PyTorch repository.\footnote{Intel-optimized PyTorch: \url{https://github.com/intel/pytorch}}

\subsubsection{I/O optimization}
I/O is challenging in many deep learning workloads partly due to random access patterns, such as those induced by shuffling, disturbing any pre-determined access order. In order to reduce the number of random access I/O operations, we developed a parallel trace sorting algorithm and pre-sorted the 15M traces according to trace type (Section \ref{sec:single_node_improvements_etalumis}). We further grouped the small trace files into larger files, going from 750 files with 20k traces per file to 150 files with 100k traces per file. With this grouping and sorting, we ensured that I/O requests follow a sequential access onto a contiguous file region which further improved the I/O performance. Metadata operations are also costly, so we enhanced the Python shelve module's file open/close performance with a caching mechanism, which allows concurrent access from different ranks to the same file.

Specific to our PPL setting, training data consists of execution traces that have a complex hierarchy, with each trace file containing many trace objects that consist of variable sequences of sample objects representing random number draws, which further contain variable length tensors, strings, integers, booleans, and other generic Python objects. PyTorch serialization with pickle is used to handle the complex trace data structure, but the pickle and unpickle overhead are very high. We developed a ``pruning'' function to shrink the data by removing non-necessary structures. We also designed a dictionary of simulator addresses $A_t$, which accumulates the fairly long address strings and assigns shorthand IDs that are used in serialization. This brought a 40\% memory consumption reduction as well as large disk space saving.

For distributed training, we developed distributed minibatch sampler and dataset classes conforming to the PyTorch training API. The sampler first splits the sorted trace indices into minibatch-sized chunks, so that all traces in each minibatch are highly likely to be of the same type, then optionally groups these chunks into several buckets (Section~\ref{sec:loadbalance}). Within each bucket, the chunks are assigned with a round-robin algorithm to different ranks, such that each rank has roughly same distribution of workload. The distributed sampler enables us to scale the training at 1,024 nodes.

The sorting of traces and their grouping into minibatch chunks significantly improves the training speed (up to 50$\times$ in our experiments) by enabling all traces in a minibatch to be propagated through the NN in the same forward execution, in other words, decreasing the need for ``sub-minibatching'' (Section~\ref{sec:single_node_improvements_etalumis}). This sorting and chunking scheme generates minibatches that predominantly contain a single trace type. However, the minibatches used at each iteration are sampled randomly without replacement from different regions of the sorted training set, and therefore contain different trace types, resulting in a gradient unbiased in expectation during any given epoch.

In our initial profiling, the cost of I/O was more than 50\% of total run time. With these data re-structuring and parallel I/O optimizations, we reduced the I/O to less than 5\%, achieving 10x speedup at different scales.

\subsubsection{Distributed improvements to PyTorch MPI CPU code}
PyTorch has a torch.distributed backend,\footnote{https://pytorch.org/docs/stable/distributed.html} which allows scalable distributed training with high performance on both CPU and GPU clusters. Etalumis uses the MPI backend as appropriate for the synchronous SGD setting that we implement (Algorithm~\ref{alg:distributed_training}) and the HPC machines we utilize (Section~\ref{sec:systems_and_software}). We have made various improvements to this backend to enable the large-scale distributed training on CPU systems required for this project. The call
{\tt torch.distributed.all\_reduce} is used to combine the gradient tensors for all distributed MPI ranks. In Etalumis, the set of non-null gradient tensors differs for each rank and is a small fraction of the total set of tensors. Therefore we first perform an allreduce to obtain a map of all the tensors that are present on all ranks; then we create a list of the tensors, filling in the ones that are not present on our rank with zero; finally, we reduce all of the gradient tensors in the list. PyTorch {\tt all\_reduce} does
not take a list of tensors so normally a list comprehension is used, but this results in
one call to {\tt MPI\_Allreduce} for each tensor. We modified PyTorch {\tt all\_reduce} to
accept a list of tensors. Then, in the PyTorch C++ code for allreduce, we concatenate small tensors into a buffer, call {\tt MPI\_Allreduce} on the buffer, and copy the results back to the original tensor. This eliminates almost all the allreduce latency and makes the communication bandwidth-bound.

We found that changing Etalumis to reduce only the non-null gradients gives a 4x improvement in allreduce time. Tensor concatenation improves overall performance by an additional 4\% on one node which increases as nodes are added. With these improvements, the load balance effects discussed in Sections  \ref{sec:scalingperf} and \ref{sec:loadbalance} are dominant and so are our primary focus of further distributed optimizations. Other future work could include performing the above steps for each backward layer with an asynchronous allreduce to overlap the communications for the previous layer with the computation for the current layer.

\section{Systems and Software}
\label{sec:systems_and_software}
\subsection{Cori}
We use the ``data'' partition of the Cori system at the National Energy Research Scientific Computing Center (NERSC) at Lawrence Berkeley National Laboratory. Cori is a Cray XC40 system, and the data partition features 2,388 nodes. Each node has two sockets and each socket is populated with a 16-core 2.3 GHz Intel\textsuperscript{\textregistered} Xeon\textsuperscript{\textregistered} E5-2698 v3 CPU (referred to as HSW from now on), with peak single-precision (SP) performance of 1.2 Tflop/s and 128 GB of DDR4-2133 DRAM. Nodes are connected via the Cray Aries low-latency, high-bandwidth interconnect utilizing the dragonfly topology. In addition, the Cori system contains 288 Cray DataWarp nodes (also known as the “Burst Buffer”) which house the input datasets for the Cori experiments presented here. Each DataWarp node contains 2 × 3.2 TB SSDs, giving a system total of 1.8 PB of SSD storage, with up to 1.7 TB/sec read/write performance and over 28M IOP/s. Cori also has a Sonnexion 2000 Lustre filesystem, which consists of 248 Object Storage Targets (OSTs) and 10,168 disks, giving nearly 30 PB of storage and a maximum of 700 GB/sec IO performance. This filesystem is used for output files (networks and logs) for Cori experiments and both input and output for the Edison experiments.
\subsection{Edison}
We also make use of the Edison system at NERSC. Edison is a Cray XC30 system with 5,586 nodes. Each node has two sockets, each socket is populated with a 12-core 2.4 GHz Intel\textsuperscript{\textregistered} Xeon\textsuperscript{\textregistered} E5-2695 v2 CPU (referred to as IVB from now on), with peak performance of 460.8 SP Gflop/s, and 64 GB DDR3-1866 memory. Edison mounts the Cori Lustre filesystem described above.
\subsection{Diamond cluster}
In order to evaluate and improve the performance on newer Intel\textsuperscript{\textregistered} processors we make use of the Diamond cluster, a small heterogeneous cluster maintained by Intel Corporation. The interconnect uses Intel\textsuperscript{\textregistered} Omni-Path Architecture switches and host adapters. The nodes used for the results in this paper are all two socket nodes. Table \ref{table:diamond} presents the CPU models used and the three letter abbreviations used in this paper.

\begin{table}[]
    \centering
    \caption{Intel\textsuperscript{\textregistered}Xeon\textsuperscript{\textregistered} CPU models and codes}
    \label{table:diamond}
    \begin{tabularx}{\columnwidth}{Xp{8mm}}
    \toprule
	Model & Code \\
	\midrule
	 E5-2695 v2 @ 2.40GHz (12 cores/socket) & IVB \\
	 E5-2698 v3 @ 2.30GHz (16 cores/socket) & HSW \\
	 E5-2697A v4 @ 2.60GHz (16 cores/socket) & BDW \\
	 Platinum 8170 @ 2.10GHz (26 cores/socket) & SKL \\
	 Gold 6252 @ 2.10GHz (24 cores/socket) & CSL\\
	 \bottomrule
    \end{tabularx}
\end{table}

\subsection{Particle physics simulation software}
\label{sec:hepsim}

In our experiments we use Sherpa version 2.2.3, coupled to a fast 3D detector simulator that we configure to use 20x35x35 voxels. Sherpa is implemented in C++, and therefore we use the C++ front end for PPX. We couple to Sherpa by a system-wide rerouting of the calls to the random number generator, which is made easy by the existence of a third-party random number generator interface (External\_RNG) already present in Sherpa.

For this paper, in order to facilitate reproducible experiments, we run in the offline training mode and produce a sample of 15M traces that occupy 1.7 TB on disk. Generation of this 15M dataset was completed in 3 hours on 32 IVB nodes of Edison. The traces are stored using Python shelve\footnote{\url{https://docs.python.org/3/library/shelve.html}} serialization, allowing random access to all entries contained in the collection of files with 100k traces in each file. These serialized files are accessed via the Python dbm module using the gdbm backend.


\section{Experiments and Results}
\subsection{Single node performance}
\label{sec:singlenoderesults}

\begin{figure}
    \centering
    \includegraphics[width=0.8\columnwidth]{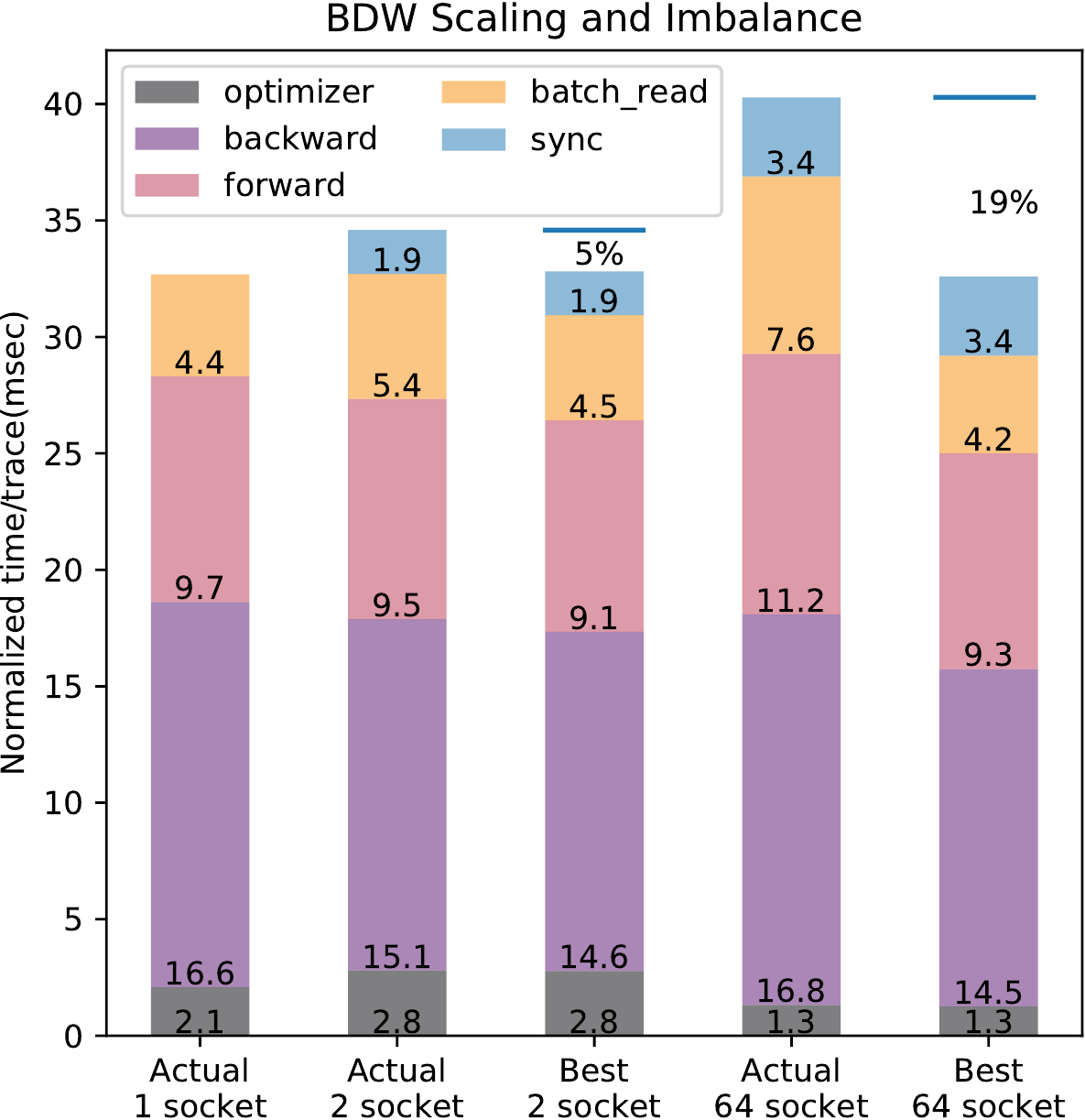}
    \caption{Actual and estimated best times for 1, 2, and 64 sockets. Horizontal bars at the top are to aid comparison between columns.}
    \label{fig:imbal}
\end{figure}

We ran single node tests with one rank per socket for one and two ranks on the IVB nodes on Edison, the HSW partition of Cori and the BDW, SKL, and CSL nodes of the Diamond cluster. Table \ref{table:throughput} shows the throughput and single socket flop rate and percentage of peak theoretical flop rate. We find that the optimizations described in Section \ref{sec:singlenodeimprove} provide an improvement of 7x on the overall single socket run throughput (measured on HSW) relative to a default PyTorch version v1.0.0 installed via the official conda channel. We achieve \textbf{430 SP Gflop/s} on a single socket of the BDW system, measured using the available hardware counters for 256-bit packed SIMD single precision operations. This includes IO and is averaged over an entire 300k trace run. This can be compared to a theoretical peak flop rate for that BDW socket of 1,331 SP Gflop/s. Flop rates for other platforms are scaled from this measurement and given in Table  \ref{table:throughput}. For further profiling we instrument the code with timers for each phase of the training (in order): minibatch read, forward, backward, and optimize. Figure \ref{fig:imbal} shows a breakdown of the time spent on a single socket after the optimizations described in Sections \ref{sec:single_node_improvements_etalumis} and \ref{sec:singlenodeimprove}.\footnote{See disclaimers section after conclusions.}

\begin{table}[]
    \centering
    \caption{Single node training throughput in traces/sec and flop rate (Gflop/s). 1-socket throughput and flop rate are for a single process while 2-socket is for 2 MPI processes on a single node. }
    \label{table:throughput}
    \begin{tabularx}{\columnwidth}{Xrrr}
    \toprule
    & 1-socket &2-socket & 1-socket  \\
	Platform & traces/s	& traces/s	&  Gflop/s (\% peak) \\
	\midrule
	IVB (Edison) & 13.9 & 25.6 & 196 (43\%) \\
	HSW (Cori) & 32.1 & 56.5 & 453 (38\%) \\
	BDW	(Diamond) &30.5	&57.8	& 430 (32\%) \\
	SKL (Diamond)	&49.9	&82.7	& 704 (20\%) \\
	CSL (Diamond) &51.1	&93.1	& 720 (22\%)\\
	\bottomrule
    \end{tabularx}
\end{table}

\subsection{Multi-node performance}
\label{sec:scalingperf}
In addition to the single socket operations, we time the two synchronization (allreduce) phases (gradient and loss). This information is recorded for each rank and each minibatch. Postprocessing finds the rank with the maximum work time (sum of the four phases mentioned in Section~\ref{sec:singlenoderesults}) and adds the times together. This gives the actual execution time. Further, we compute the \emph{average} time across ranks for each work phase for each minibatch and add those together, giving the best time assuming no load imbalance. The results are shown in Figure \ref{fig:imbal}. Comparing single socket results with 2 and 64 socket results shows the increased impact of load imbalance as nodes are added. This demonstrates a particular challenge of this project where the load on each node depends on the random minibatch of traces sampled for training on that node. The challenge and possible mitigation approaches we explored are discussed further in Section \ref{sec:loadbalance}.

\begin{figure}
  \includegraphics[width=0.49\textwidth]{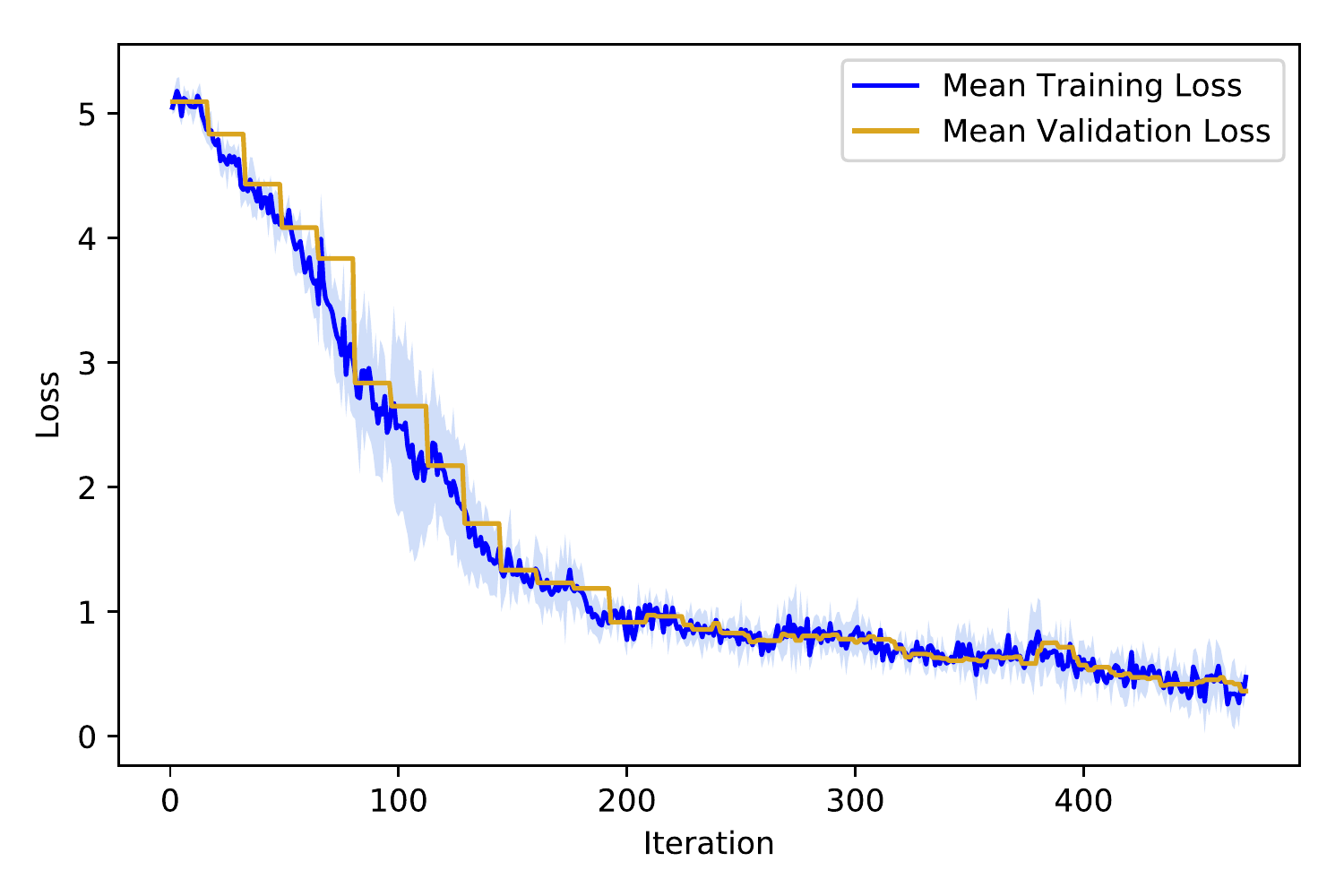}
  \caption{Mean loss and standard deviation (shaded) for five experiments with 128k minibatch size.}
  \label{fig:meanloss}
\end{figure}

\subsection{Large scale training on Cori and Edison}

In order to choose training hyperparameters, we explored global minibatch sizes of \{64, 2k, 32k, 128k\}, and learning rates in the range $\left[10^{-7}, 10^{-1}\right]$ with grid search and compared the loss values after one epoch to find the optimal learning rate for different global minibatch sizes separately. For global minibatch sizes of 2k, 32k, and 128k, we trained with both Adam and Adam-LARC optimizers and compared loss value changes as a function of iterations. For training at 1,024 nodes we choose to use 32k and 128k global minibatch sizes. For the 128k minibatch size, best convergence was found using the Adam-LARC optimizer with a polynomial decay (order=2) learning rate schedule \citep{lars-paper} that decays from an initial global learning rate of $5.70\times10^{-4}$ to final $2\times10^{-5}$ after completing 12 epochs for the dataset with 15M traces. In Figure \ref{fig:meanloss}, we show the mean and standard-deviation for five training runs with this 128k minibatch size and optimizer, demonstrating stable convergence.

\begin{figure*}
  \centering
  \includegraphics[width=0.48\textwidth]{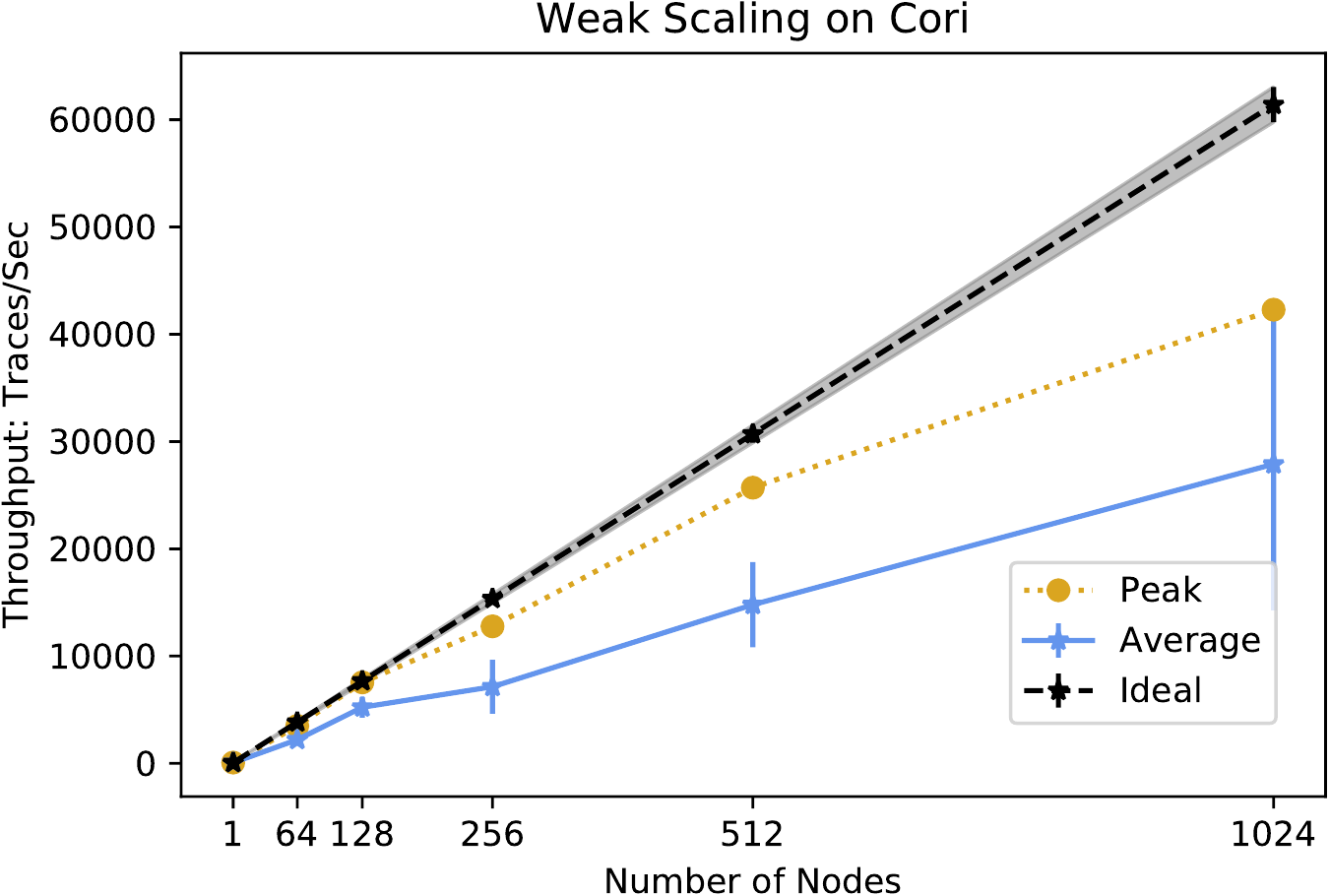}\qquad
  \includegraphics[width=0.48\textwidth]{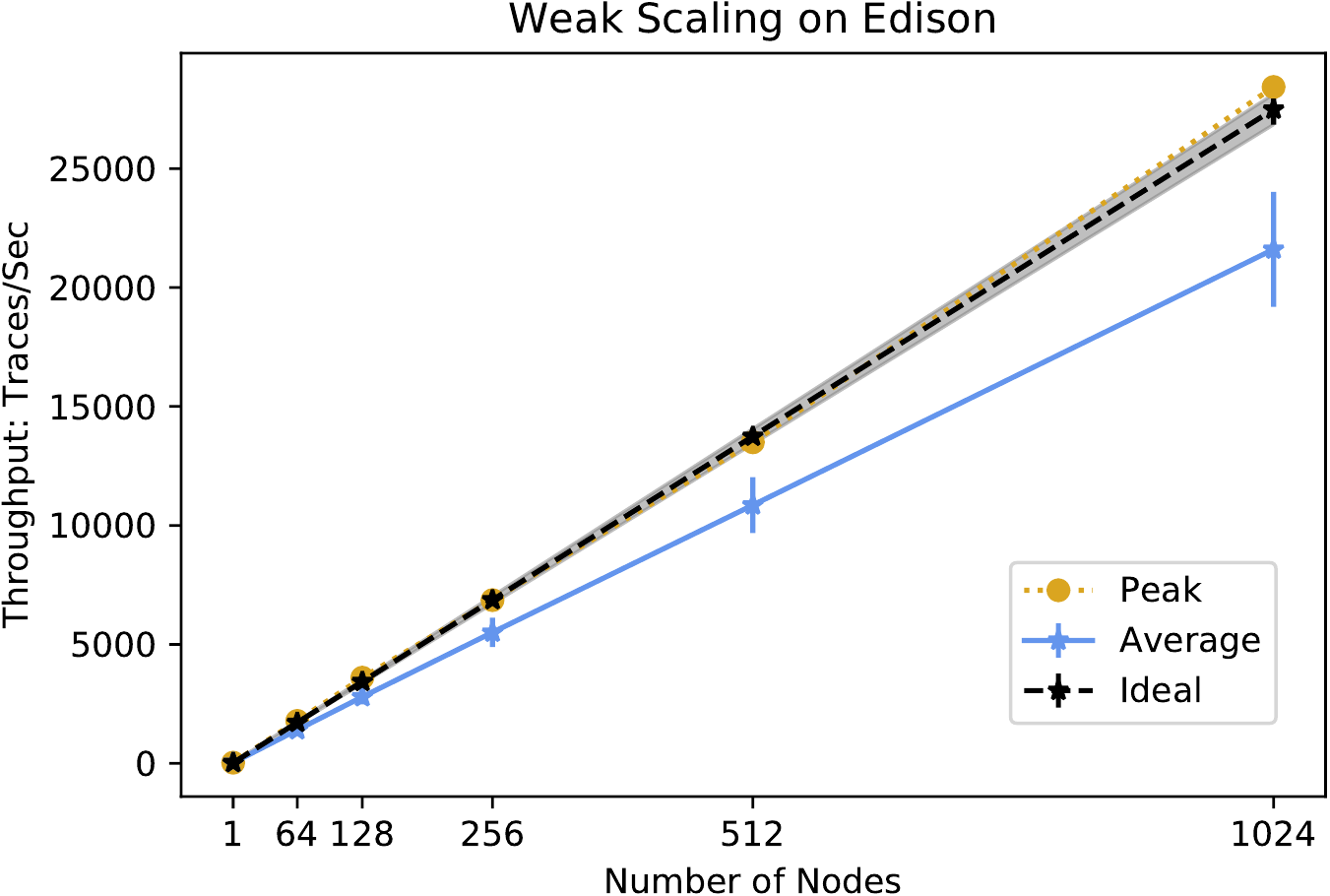}
  \vspace{3mm}
  \caption{Weak scaling on Cori and Edison systems, showing throughput for different node counts with a fixed local minibatch size of 64 traces per MPI rank with 2 ranks per node. Average (mean) over all iterations and the peak single iteration are shown.  Ideal scaling is  derived from the mean single-node rate with a shaded uncertainty from the standard deviation in that rate.}
  \label{fig:scaling}
\end{figure*}

Figure~\ref{fig:scaling} shows weak scaling results obtained for distributed training to over a thousand nodes on both the Cori and Edison systems. We use a fixed local minibatch size of 64 per rank with 2 ranks per node, and plot the mean and standard deviation throughput for each iteration in terms of traces/s (labeled ``average'' in the plot). We also show the fastest iteration (labeled ``peak''). The average scaling efficiency at 1,024 nodes is 0.79 on Edison and 0.5 on Cori. The throughput at 1,024 nodes on Cori and Edison is 28,000 and 22,000 traces/s on average, with the peak as 42,000 and 28,000 traces/s respectively. One can also see that there is some variation in this performance due to the different compute times taken to process execution traces of different length and the related load imbalance as discussed in Sections \ref{sec:scalingperf} and \ref{sec:loadbalance}. We determine the maximum sustained performance over a 10-iteration sliding window to be \textbf{450 Tflop/s} on Cori and 325 Tflop/s on Edison.\footnote{See disclaimers section after conclusions.}

\begin{figure}
  \includegraphics[width=0.49\textwidth]{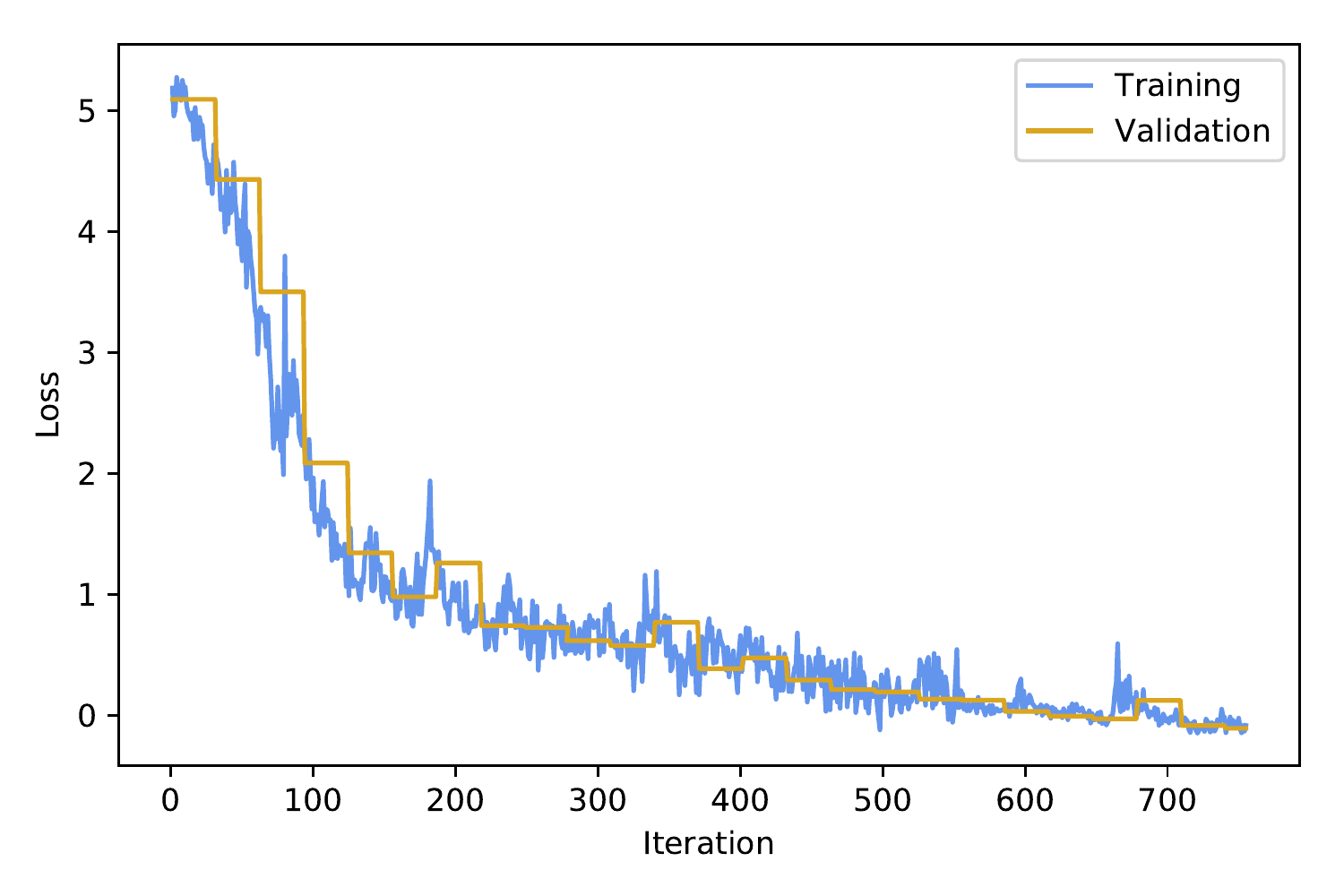}
  \caption{Training and validation loss for a 128k minibatch size experiment with the configuration described in the text run on 1,024 nodes of the Edison system.}
  \label{fig:loss}
\end{figure}

We have performed distributed training with global minibatch sizes of 32k and 128k at 1,024-node scale for extended periods to achieve convergence on both Cori and Edison systems. This is illustrated in Figure \ref{fig:loss} where we show the loss for training and validation datasets as a function of iteration for an example run on Edison.

\subsection{Inference and science results}
Using our framework and the NNs trained using distributed resources at NERSC as described previously, we perform inference on test $\tau$ observation data that has not been used for training. As the approach of applying probabilistic programming in the setting of large-scale existing simulators is completely novel, there is no direct baseline in literature that provides the full posterior in each of these variables. Therefore we use our own MCMC (RMH)-based posterior as a baseline for the validation of the IC approach. We establish the convergence of the RMH posterior by running two independent MCMC chains with different initializations and computing the the Gelman--Rubin convergence metric \citep{gelman2013bayesian} to confirm that they converge onto the same posterior distribution (Section~\ref{sec:efficient_bayesian_inference}).

\begin{figure}
  \includegraphics[trim={0.3cm 0.3cm 0.5cm 0.5cm},clip,width=0.47\textwidth]{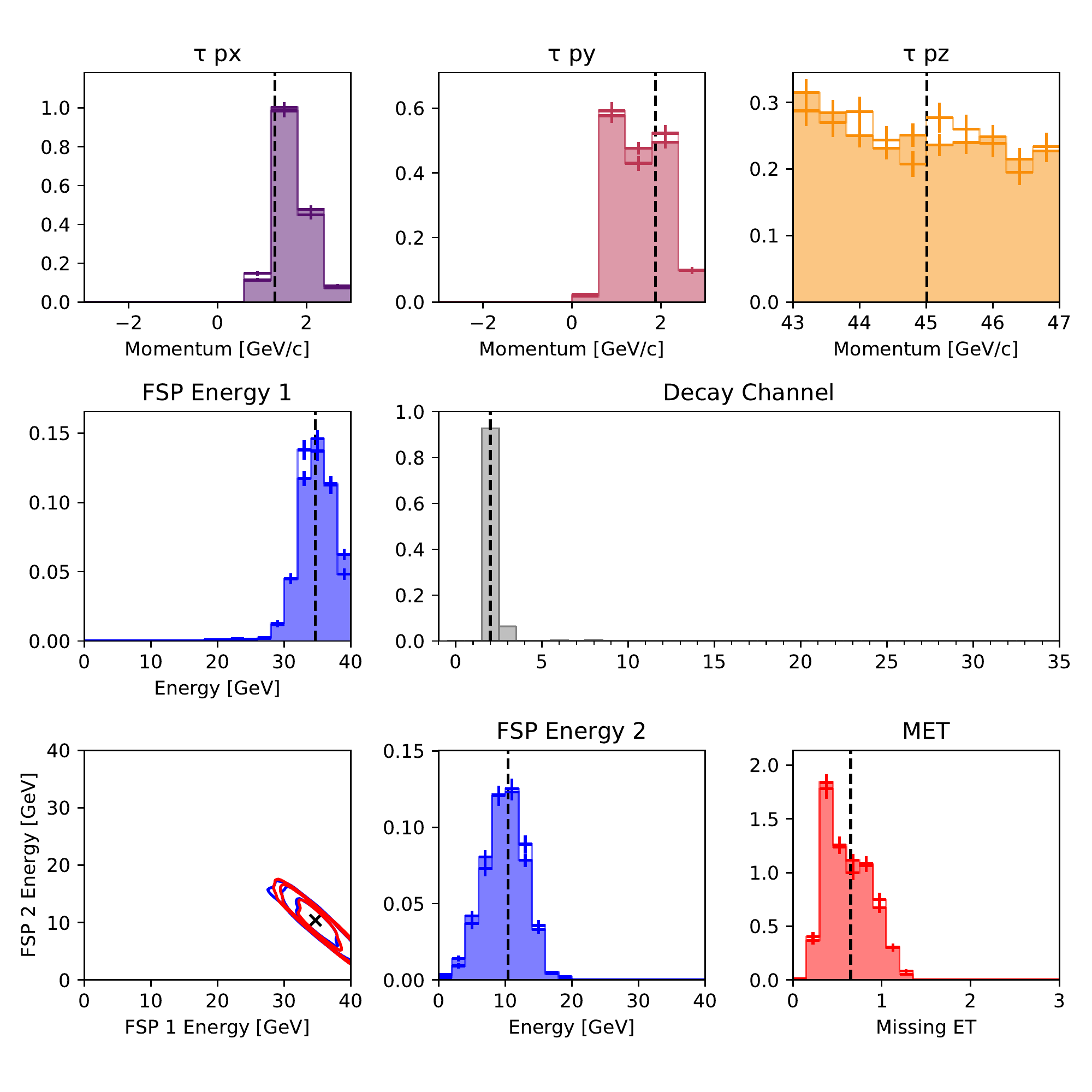}
  \vspace{3mm}
  \caption{A comparison of posterior distributions obtained by RMH (filled histograms) and IC (outline histograms) and the ground truth values (dashed vertical lines) for a test $\tau$ decay observation. We show an illustrative subset of the latent variables, including x, y and z components of the $\tau$-lepton momentum (top row), the energies of the two highest energy particles produced by the $\tau$ decay (middle left and bottom center), a contour plot showing correlation between these (bottom left), the $\tau$ decay channel ($\tau\to\pi\,\nu_{\tau}$ as mode) (middle right), and the missing transverse energy (bottom right).}
  \label{fig:physics}
\end{figure}

Figure~\ref{fig:physics} shows a comparison of inference results from the RMH and IC approaches. We show selected latent variables (addresses) that are a small subset of the more than 24k addresses that were encountered in the prior space of the Sherpa experimental setup, but are of physics interest in that they correspond to properties of the $\tau$ particle. It can be seen that there is close agreement between the RMH and IC posterior distributions validating that our network has been adequately trained. We have made various improvements to the RMH inference processing rate but this form of inference is compute intensive and takes \textbf{115 hours} on a Edison IVB node to produce the 7.68M trace result shown. The corresponding 2M trace IC result completed in \textbf{30 mins} (achieving a 230$\times$ speedup for a comparable posterior result) on 24 HSW nodes, enabled by the parallelism of IC inference.

In addition to parallelization, a significant advantage of the IC approach is that it is amortized. This means that once the proposal NN is trained for any given model, it can be readily applied to large volumes of new collision data. Moreover IC inference runs with high effective sample sizes in comparison to RMH: each sample from the IC NN is an independent sample from the proposal distribution, which approaches the true posterior distribution with more training, whereas our autocorrelation measurements in the RMH posterior indicate that a very large number of iterations are needed to get statistically independent traces (on the order of $\sim 10^5$ for the type of decay event we use as the observation). These features of IC inference combine to provide a tractable approach for fast Bayesian inference in complex models implemented by large-scale simulators.

\section{Discussion}
\label{sec:discussion}
The dynamic NN architecture employed for this project has presented a number of unique challenges for distributed training, which we covered in Section~\ref{sec:scaling_nn}. Our innovations proved successful in enabling the training of this architecture at scale, and in this section we capture some of the lessons learned and unresolved issues encountered in our experiments.

\subsection{Time to solution: trade-off between throughput and convergence}

\subsubsection{Increasing effective local minibatch size}
\label{sec:discusslocalbatch}
As mentioned in Section \ref{sec:single_node_improvements_etalumis}, the distributed SGD scheme given in Algorithms \ref{alg:minibatch_loss} and \ref{alg:distributed_training} uses random traces sampled from the simulator, and can suffer from slow training throughput if computation cannot be efficiently parallelized across the full minibatch due to the presence of different trace types. Therefore, we explored the following methods to improve effective minibatch size: sorting the traces before batching, online batching of the same trace types together, and multi-bucketing. Each of these methods can improve throughput, but risk introducing bias into SGD training or increasing the number of iterations to converge, so we compare the wall clock time for convergence to determine the relative trade-off.  The impact of multi-bucketing is described and discussed in section \ref{sec:loadbalance} below. Sorting and batching traces from the same trace type together offers considerable throughput increase with a relatively small impact on convergence, when combined with shuffling of minibatches for randomness, so we used these throughput optimizations for the results in this paper.

\subsubsection{Choice of optimizers, learning rate scaling and scheduling for convergence}
\label{sec:discusoptlr}
There is considerable recent literature on techniques for distributed and large-minibatch training, including scaling learning rates  with number of nodes \citep{Goyal}, alternative optimizers for large-minibatch-size training, and learning rate decay \citep{Adam,larc-paper,lars-paper}. The network employed in this project presents a very different use-case to those considered in literature, prompting us to document our experiences here. For learning rate scaling with node count, we found sub-sqrt learning rate scaling works better than linear for an Adam-based optimizer \citep{Adam}. We also compared Adam with the newer Adam-LARC optimizer \citep{larc-paper, lars-paper} for convergence performance and found the Adam-LARC optimizer to perform better for the very large global minibatch size 128k in our case. For smaller global minibatch sizes of 32K or lower, both plain Adam and Adam-LARC performed equally well. Finally, for the learning rate decay scheduler, we explored the following options: no decay, multi-step decay (per epoch), and polynomial decay of order 1 or 2 (calculated per iteration) \citep{lars-paper}. We found that learning-rate decay can improve training performance and polynomial decay of order 2 provided the most effective schedule.

\subsection{Load balancing}
\label{sec:loadbalance}
As indicated in Section \ref{sec:scalingperf}, our work involves distinct scaling challenges due to the variation in compute time depending on execution trace length, address-dependent proposal and embedding layers, and representation of trace types inside each minibatch. These factors contribute to load imbalance. The trace length variation bears similarity to varying sequence lengths in NMT; however, unlike that case it is not possible to truncate the execution traces, and padding would introduce a cost to overall number of operations. 

To resolve this load imbalance issue, we have explored a number of options, building on those from NMT, including a \emph{multi-bucketing} scheme and a novel \emph{dynamic batching} approach.

In \emph{multi-bucketing} \citep{bengio2009curriculum,khomenko2016accelerating,doetsch2017comprehensive}, traces are grouped into several buckets based on lengths, and every global minibatch is solely taken from a randomly picked bucket for every iteration. Multi-bucketing not only helps to balance the load among ranks for the same iteration, but also increases the effective minibatch size as traces from the same trace type have a higher chance to be in the same minibatch than in the non-bucketing case, achieving higher throughput. For a local minibatch-size of 16 with 10 buckets we measured throughput increases in the range of 30--60\% at 128--256 node scale on Cori. However, our current multi-bucketing implementation does multiple updates in the same bucket continuously. When this implementation is used together with batching the traces from the same trace type together (as discussed above in Section~\ref{sec:discusslocalbatch}), it negatively impacts convergence behavior. We believe this to be due to the fact that training on each specific bucket for multiple updates introduces over-fitting onto that specific subset of networks so moving to a new bucket for multiple updates causes  information of the progress made with previous buckets to be lost. As such we did not employ this configuration for the results reported in this paper.

With \emph{dynamic batching}, we replaced the requirement of fixed minibatch size per rank with a desired number of ``tokens'' per rank (where a token is a unit of random number draws in each trace), so that we can, for instance, allocate many short traces (with smaller number of tokens each) for one rank but only a few long traces for another rank, in order to balance the load for the LSTM network due to length variation. While an equal-token approach has been used in NMT, this did not offer throughput gains for our model, which has an additional 3DCNN component in which the compute time depends on the number of traces within the local minibatch, so if dynamic batching only considers total tokens per rank for the LSTM it can negatively impact the 3DCNN load.

Through these experiments we found that our current optimal throughput and convergence performance came from not employing these load-balancing schemes although we intend to explore modifications to these schemes as ongoing work.

\section{Science implications and outlook}
We have provided a common interface to connect PPLs with simulators written in arbitrary code in a broad range of programming languages. This opens up possibilities for future work in all applied fields where simulators are used to model real-world systems, including epidemiology modeling such as disease transmission and prevention models \citep{smith_maire_ross_2008}, autonomous vehicle and reinforcement learning environments \citep{Dosovitskiy17}, cosmology \citep{akeret2015approximate}, and climate science \citep{sterman2012climate}. In particular, the ability to control existing simulators at scale and to generate interpretable posteriors is relevant to scientific domains where interpretability in model inference is critical.

We have demonstrated both MCMC- and IC-based inference of detector data originating from $\tau$-decays simulated with the Sherpa Monte Carlo generator at scale. This offers, for the first time, the potential of Bayesian inference on the full latent structure of the large numbers of collision events produced at accelerators such as the LHC, enabling deep interpretation of observed events. For instance, ambiguity in the decay of a particle can be related exactly to the physics processes in the simulator that would give rise to that ambiguity. In order to fully realize this potential, future work will expand this framework to more complex particle decays (such as the Higgs decay to $\tau$ leptons) and incorporate a more detailed detector simulation (e.g., Geant4 \citep{agostinelli2003geant4}).  We will demonstrate this on a full LHC physics analysis, reproducing the efficiency of point-estimates, together with the full posterior and intpretability, so that this can be exploited for discovery of new fundamental physics. 

The IC objective is designed so that the NN proposal $q(\mathbf{x}\vert\mathbf{y})$ approximates the posterior $p(\mathbf{x}\vert\mathbf{y})$ asymptotically closely with more training. This costly training phase needs to be done \emph{only once} for a given simulator-based model, giving us a NN that can provide samples from the model posterior in parallel for any new observed data. In this setting where have a fast, amortized $q(\mathbf{x}\vert\mathbf{y})\approx p(\mathbf{x}\vert\mathbf{y})$, our ultimate goal is to add the machinery of Bayesian inference to the toolbox for critical tasks such as triggering \citep{Aaboud:2016leb} and event reconstruction by conditioning on potentially interesting events (e.g., $q(\mathrm{ParticleType}\vert \cdot)\ge\epsilon$). Recent activity exploring the use of FPGAs for NN inference for particle physics \citep{Duarte:2018ite} will help implementation of these approaches, and HPC systems will be crucial in the training and inference phases of such frameworks.

\section{Conclusions}
Inference in simulator-based models remains a challenging problem with potential impact across many disciplines \citep{papamakarios2016fast,brehmer2018mining}. In this paper we present the first probabilistic programming implementation capable of controlling existing simulators and running at large-scale on HPC platforms. Through the PPX protocol, our framework successfully couples with large-scale scientific simulators leveraging thousands of lines of existing simulation code encoding domain-expert knowledge. To perform efficient inference we make use of the inference compilation technique, and we train a dynamic neural network involving LSTM and 3DCNN components, with a large global minibatch size of 128k. IC inference achieved a 230$\times$ speedup compared with the MCMC baseline. We optimize the popular PyTorch framework to achieve a significant single-socket speedup for our network and 20--43\% of peak theoretical flop rate on a range of current CPUs.\footnote{See disclaimers section after conclusions.} We augment and develop PyTorch's MPI implementation to run it at the unprecedented scale of 1,024 nodes (32,768 and 24,576 cores) of the Cori and Edison supercomputers with a sustained flop rate of 0.45 Pflop/s. We demonstrate we can successfully train this network to convergence at these large scales, and use this to perform efficient inference on LHC collision events. The developments described here open the door for exploiting HPC resources and existing detailed scientific simulators to perform rapid Bayesian inference in very complex scientific settings.

\section*{Disclaimers}
\label{sec:disclaimers}
\footnotesize
Software and workloads used in performance tests may have been optimized for performance only on Intel microprocessors. 

Performance tests, such as SYSmark and MobileMark, are measured using specific computer systems, components, software, operations and functions. Any change to any of those factors may cause the results to vary. You should consult other information and performance tests to assist you in fully evaluating your contemplated purchases, including the performance of that product when combined with other products. 

For more complete information visit \url{www.intel.com/benchmarks}.

Performance results are based on testing as of March 22 and March 28, 2019 and may not reflect all publicly available security updates. See configuration disclosure for details. No product or component can be absolutely secure. 

Configurations: Testing on Cori and Edison (see \S\ref{sec:singlenoderesults}) was performed by NERSC. Testing on Diamond cluster was performed by Intel.

Intel technologies' features and benefits depend on system configuration and may require enabled hardware, software or service activation. Performance varies depending on system configuration. Check with your system manufacturer or retailer or learn more at \url{www.intel.com}.

Intel does not control or audit third-party benchmark data or the web sites referenced in this document. You should visit the referenced web site and confirm whether referenced data are accurate. 

Intel, VTune, Xeon are trademarks of Intel Corporation or its subsidiaries in the U.S. and/or other countries.

\begin{acks}
The authors would like to acknowledge valuable discussions with Thorsten Kurth on scaling aspects, Quincey Koziol on I/O; Steve Farrell on NERSC PyTorch, Holger Schulz on Sherpa, and Xiaoming Cui, from Intel AIPG team, on NMT. This research used resources of the National Energy Research Scientific Computing Center (NERSC), a U.S. Department of Energy Office of Science User Facility operated under Contract No. DE-AC02-05CH11231. This work was partially supported by the NERSC Big Data Center; we acknowledge Intel for their funding support. KC, LH, and GL were supported by the National Science Foundation under the awards ACI-1450310. Additionally, KC was supported by the National Science Foundation award OAC-1836650. BGH is supported by the EPRSC Autonomous Intelligent Machines and Systems grant. AGB and PT are supported by EPSRC/MURI grant EP/N019474/1 and AGB is also supported by Lawrence Berkeley National Lab. FW is supported by DARPA D3M, under Cooperative Agreement FA8750-17-2-0093, Intel under its LBNL NERSC Big Data Center, and an NSERC Discovery grant.
\end{acks}

\bibliographystyle{ACM-Reference-Format}
\bibliography{main}

\end{document}